\RequirePackage[T1]{fontenc}
\documentclass[letterpaper,10pt,conference]{ieeeconf}
\IEEEoverridecommandlockouts
\usepackage{newtxtext}
\usepackage{cite}
\usepackage{amsmath,amssymb,bm}
\usepackage{graphicx,placeins,float,algorithmic,tikz,url,balance,array}
\let\labelindent\relax
\usepackage{enumitem}
\usepackage[hidelinks,draft]{hyperref}
\usetikzlibrary{arrows.meta,positioning}
\newcounter{lacealgorithm}
\newcommand{\scorecell}[2]{\shortstack[r]{#1\\{\scriptsize$\pm #2$}}}
\interdisplaylinepenalty=2500 \makeatletter \g@addto@macro\normalsize{\setlength{\abovedisplayskip}{5pt plus 2pt minus 2pt}\setlength{\belowdisplayskip}{5pt plus 2pt minus 2pt}\setlength{\abovedisplayshortskip}{0pt plus 2pt}\setlength{\belowdisplayshortskip}{3pt plus 2pt minus 1pt}}\makeatother
\setlist[itemize]{leftmargin=1.35em,itemsep=1pt,topsep=2pt,parsep=0pt}
\hypersetup{pdfauthor={Yuhan Wen, Jiawei Wang, Qixuan Zhang, Calvin Zhang, Yusen Qin, Lan Xu},pdftitle={LACE-CRAFT: Robot Co-Design with Actor Inheritance and Blackboard Collaboration}}
\ifdefined
\fi
\begin{document}
\bstctlcite{lace_bibliography_layout}
\title{\LARGE\bfseries LACE--CRAFT: Robot Co-Design with Actor Inheritance and Blackboard Collaboration}

\author{\large Yuhan Wen$^{1,2}$\quad Jiawei Wang$^{1,3}$\quad Qixuan Zhang$^{1,2}$\quad Calvin Zhang$^{5}$\quad
Yusen Qin$^{4}$\quad Lan Xu$^{1,2,*}$\\[3pt]{$^{1}$Deemos Corporation \quad $^{2}$ShanghaiTech University \quad $^{3}$University of Cambridge}\\[2pt]{$^{4}$D-Robotics \quad $^{5}$Independent Researcher}\thanks{\raggedright \textsuperscript{*}Corresponding author: Lan Xu (\texttt{xulan1@shanghaitech.edu.cn}).}}

\IEEEoverridecommandlockouts
\IEEEaftertitletext{%
\vspace{-12pt}
\begin{center}
\includegraphics[width=0.96\textwidth]{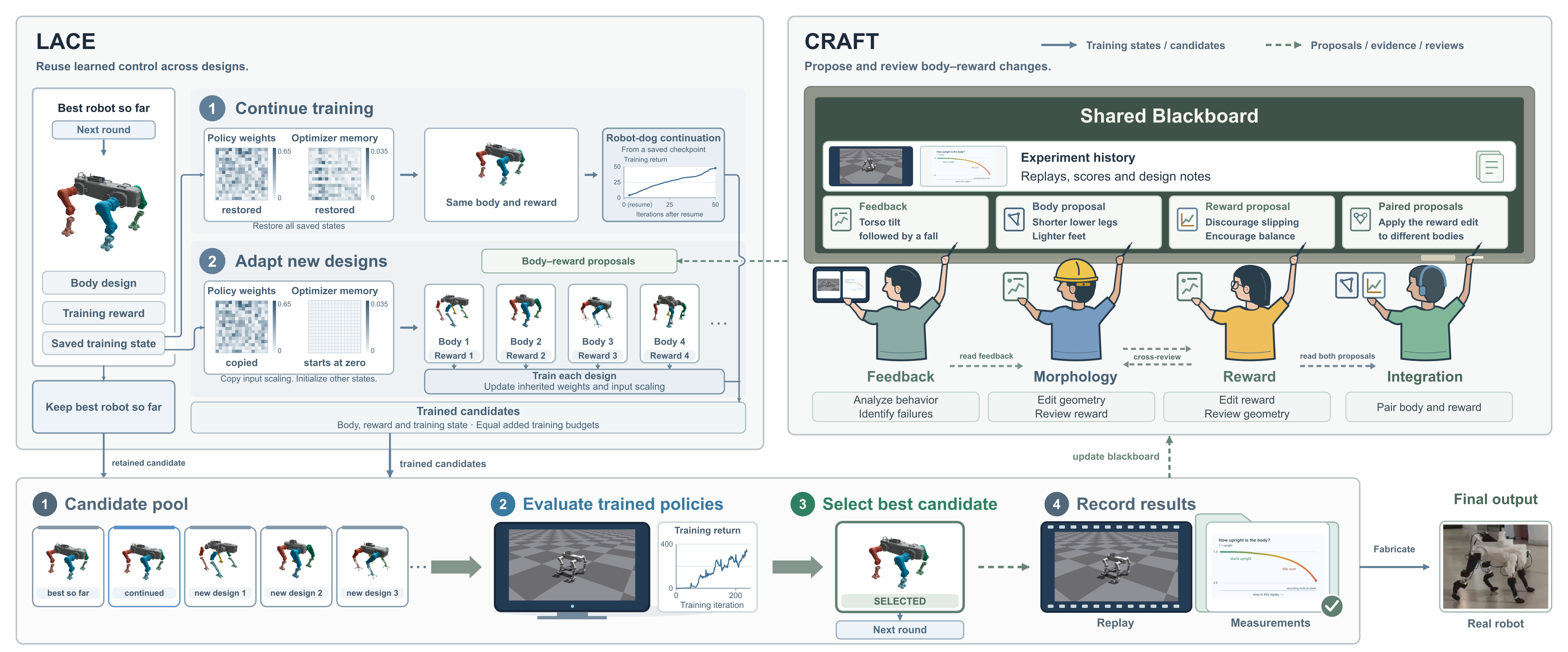}\\[-1mm]
\vspace{1mm}
\refstepcounter{figure}\label{fig:lace-craft-overview}%
\parbox{0.96\textwidth}{\footnotesize\raggedright Fig.~\thefigure. LACE--CRAFT combines policy reuse with body--reward co-design. CRAFT reviews and pairs body and reward proposals using shared experiment records. Both LACE branches independently start from the same round-start state. Continuation restores the full training state, while adaptation inherits policy weights and input normalization and initializes other states anew. With equal additional environment-step budgets, both branches produce candidates scored by a fixed task metric and compared against the retained incumbent's stored score. The winner initializes the next round, and recorded results update the blackboard. Solid arrows carry states and candidates. Dashed arrows carry proposals, reviews and evidence. The bridge indicates no connection. Outside the iterative loop, the final design is fabricated as a real robot.}
\end{center}
\vspace{-0.5\baselineskip}
}
\maketitle
\thispagestyle{empty}
\pagestyle{empty}
\raggedbottom

\begin{abstract}
Robot co-design couples morphology search with policy learning, yet training every new design from scratch discards acquired control experience. We present LACE--CRAFT, which compares continued learning on the current robot with policy adaptation to new morphology--reward pairs. LACE resumes the incumbent's full learning state and initializes compatible challengers with its actor parameters and observation statistics. A fixed task metric selects among both branches and the frozen incumbent. CRAFT coordinates Feedback, Morphology, Reward, and Integration roles through shared experimental records and behavioral replays to generate and cross-review paired proposals. A generative extension converts generated meshes into editable articulated models with configured joints, actuator interfaces, and consistently updated simulation assets. Across five locomotion benchmarks, mean scores over three evaluation seeds are 6.4--91.9\% higher than D2C. Both methods train 30 new morphology--reward pairs over five rounds; LACE additionally uses four continuation training units. Five-task ablations examine policy inheritance and replay-derived feedback. A fabricated prototype demonstrates indoor walking and illustrates the geometry-to-hardware workflow.
\end{abstract}

\section{Introduction}
Robot morphology determines the dynamics a controller must exploit, while the training reward shapes the behavior it learns. Co-design therefore couples body design, reward design, and policy optimization~\cite{luck2020data,fang2025robomore}. Large language models (LLMs) provide flexible operators for this search: Eureka generates executable rewards from training feedback~\cite{ma2024eureka}, and Debate2Create (D2C) uses multi-agent debate to refine morphology--reward pairs before simulation-based evaluation~\cite{qiu2026debatecreate}.

The search must also decide how to use the control experience acquired during candidate evaluation. In our D2C reproduction, design records guide subsequent proposals, while each new policy is trained from scratch. Prior co-design methods already reuse experience or transfer controller parameters across morphologies~\cite{luck2020data,wang2018neural}. We focus on how to select between further training the current design and adapting its policy to a new body and reward. A learned policy can provide a useful initialization after a design change, whereas value estimates and stored transitions depend on the dynamics and reward under which they were obtained.

LACE--CRAFT addresses this decision through two competing training branches (Fig.~\ref{fig:lace-craft-overview}). LACE initializes new designs with the incumbent's actor parameters and observation statistics, while resetting other learner components. A continuation branch resumes the incumbent's complete learning state on its unchanged morphology and reward. A fixed external task metric compares both branches with the frozen incumbent, allowing search to select either further learning on the current design or adaptation to a new design.

CRAFT uses the resulting experimental evidence to guide subsequent proposals. Its shared blackboard links tested morphology--reward pairs to checkpoints, scores, and behavioral replays. Feedback proposes diagnoses and improvements; Morphology and Reward independently draft and then cross-review changes; Integration constructs complete paired candidates. New evaluations update the blackboard for the next round, preserving the association between each result and the robot that produced it.

We further extend the morphology representation from parameterized benchmark bodies to generated meshes. Following the geometry-to-hardware approach of Text2Robot~\cite{ringel2025text2robot}, we process generated shapes into editable links, configured joints, and actuator interfaces. Geometry edits update the articulated model and its simulation assets while preserving LACE's policy interface. Alongside the five-task benchmark study, we document this processing workflow and a fabricated walking prototype. Our contributions are:
\begin{itemize}
\item \textbf{Continuation-aware co-design.} LACE compares full-state continuation of the incumbent with actor-initialized morphology--reward challengers under a fixed task metric, while retaining the incumbent as a selection option.
\item \textbf{Evidence-informed proposal generation.} CRAFT coordinates four specialist roles through shared experiment records, diagnostic feedback, reciprocal morphology--reward review, and integration.
\item \textbf{A generative morphology extension.} Editable links, configured joints, and actuator interfaces connect generated geometry to the co-design representation; processing records and a walking prototype illustrate its use.
\end{itemize}

\section{Related Work}
\paragraph{Morphology--control co-design}
Evolutionary co-design spans simulated creatures~\cite{sims2023evolving} and fabricated robots~\cite{lipson2000automatic}. Later approaches optimize body parameters or design distributions alongside controllers~\cite{ha2019reinforcement,schaff2019jointly,luck2020data}. RoboGrammar constrains structural search~\cite{zhao2020robogrammar}, while Transform2Act combines design and control~\cite{yuan2022transformact}. DERL couples evolution with lifetime learning~\cite{gupta2021embodied}. Morphological innovation protection allows controllers to adapt before new bodies face selection~\cite{cheney2018scalable}, highlighting the interaction between design changes and learning progress.

\paragraph{Controller transfer}
Hardware conditioning~\cite{chen2018hardware}, graph policies~\cite{wang2018nervenet}, and shared actuator modules~\cite{huang2020one} support cross-morphology control. Amorpheus examines morphological graph biases~\cite{kurin2021my}, whereas MetaMorph and ModuMorph use Transformer-based conditioning~\cite{gupta2022metamorph,xiong2023universal}. N-LIMB jointly learns designs and a universal controller~\cite{schaff2022n}. Neural Graph Evolution inherits parental policies~\cite{wang2018neural}, and controller archives retrieve policies for related body plans~\cite{le2022morpho}. LACE compares compatible actor transfer with continued learning of the retained robot.

\paragraph{Language-guided design}
Language models provide proxy reward signals~\cite{kwon2023reward}, synthesize dense reward programs~\cite{xie2024text2reward}, and translate instructions into control objectives~\cite{pmlr-v229-yu23a}. Code as Policies instead generates executable policy programs~\cite{liang2023code}. Eureka iteratively refines rewards using training feedback~\cite{ma2024eureka}. RoboMorph explores morphology~\cite{qiu2026robomorph}, while RoboMoRe and D2C jointly propose morphology--reward changes~\cite{fang2025robomore,qiu2026debatecreate}.

\paragraph{Agent collaboration}
ReAct interleaves reasoning and actions~\cite{yao2023react}. Reflexion retains verbal feedback~\cite{shinn2023reflexion}, and Self-Refine revises outputs through self-critique~\cite{madaan2023self}. AutoGen supports configurable conversations~\cite{wu2024autogen}, MetaGPT structures role-specific artifacts~\cite{hong2024metagpt}, and CAMEL organizes role-conditioned cooperation~\cite{li2023camel}. Multiagent debate enables reciprocal critique~\cite{pmlr-v235-du24e}. CRAFT combines role specialization with a blackboard-style shared record~\cite{hayes1985blackboard}, grounding proposals in experimental evidence.

\paragraph{Generated geometry and deployment}
CLAY generates static 3D assets~\cite{zhang2024clay}, BANG decomposes parts~\cite{zhang2025bang}, and Text2Robot connects generated geometry to fabrication~\cite{ringel2025text2robot}. RoboGen generates simulation tasks and training supervision~\cite{pmlr-v235-wang24cc}. Our extension exposes generated bodies as editable articulated morphologies. GPU simulation~\cite{makoviychuk2021isaac} and parallel locomotion training~\cite{rudin2022learning} accelerate policy optimization. Actuator modeling and robust training support physical quadruped deployment~\cite{Tan-RSS-18,hwangbo2019learning}, while DrEureka automates reward and domain-randomization proposals~\cite{Ma-RSS-24}.

\section{LACE--CRAFT Method}

\subsection{Problem Formulation}

Let $\mathcal M$, $\mathcal Z$, and $\mathcal R$ denote the morphology, training-state, and training-reward spaces. The training state $z$ contains the policy and the learner variables needed to resume PPO~\cite{schulman2017proximal} or SAC~\cite{haarnoja2018soft}, as listed in Table~\ref{tab:state-transfer}. We distinguish the robot-and-learning state $x$ from the search state $y$, which also retains its training reward:

\begin{equation}
\begin{gathered}
x=(m,z),\qquad y=(x,r),\\
(m,z,r)\in\mathcal M\times\mathcal Z\times\mathcal R.
\end{gathered}
\label{eq:state}
\end{equation}

A reward $r$ and transition budget $B\in\mathbb N^{+}$ specify the learning update $\mathcal U_B^r$. On morphology $m$, it performs reinforcement learning for $B$ environment transitions, summed across parallel environments:

\begin{equation}
z^{+}=\mathcal U_B^{r}(m;z).
\label{eq:update}
\end{equation}
The updated state is $x^+=(m,z^+)$. The projection $\Pi$ extracts the executable observation-to-action policy, including its input normalization. Deployment uses $(m,\Pi(z))$; auxiliary training variables remain in $z$ for subsequent learning.

For task $\tau$, let $g_\tau(s_{t+1})$ be the fixed evaluation signal after action $a_t$ and $T_\tau$ the evaluation horizon. Following external task-fitness evaluation in reward design and co-design~\cite{ma2024eureka,qiu2026debatecreate}, we define
\begin{equation}
S_\tau(m,\pi)=\mathbb E_{\rho\sim p(\cdot\mid m,\pi)}\left[\sum_{t=0}^{T_\tau-1}g_\tau(s_{t+1})\right].
\label{eq:task-score}
\end{equation}
Here $\rho$ is a rollout drawn from the trajectory distribution $p(\cdot\mid m,\pi)$ under morphology $m$ and policy $\pi$; $s_t$ and $a_t$ are its state and action at timestep $t$. Contributions after termination are zero. The signal $g_\tau$ and horizon $T_\tau$ are fixed across candidates, independently of their training rewards. Following the D2C benchmark, we accumulate distance from the origin over each episode using the root-position coordinates exposed by each task environment. This metric differs from path length and training return. Selection uses the finite-evaluation estimate
\begin{equation}
\widehat S_\tau(m,\pi)=\frac{1}{N_{\rm eval}}\sum_{\ell=1}^{N_{\rm eval}}\sum_{t=0}^{T_\tau-1}g_\tau(s_{t+1}^{(\ell)}),
\label{eq:empirical-score}
\end{equation}
where $N_{\rm eval}$ is the number of evaluation episodes and $\ell$ indexes an episode.

Let $\mathcal X_\tau\subseteq\mathcal M\times\mathcal Z$ contain the admissible morphology--state pairs with a compatible policy interface. The co-design objective is
\begin{equation}
\max_{(m,z)\in\mathcal X_\tau}
S_\tau\bigl(m,\Pi(z)\bigr).
\label{eq:objective}
\end{equation}

The reward $r$ guides the finite-budget learning update, while $S_\tau$ evaluates the resulting robot. LACE--CRAFT searches through agent-proposed morphology--reward pairs and RL updates, retaining the complete state needed to continue the selected candidate.

\subsection{LACE: Learning-State-Aware Continuation Evolution}

After initialization of $y_0$, LACE runs for at most $K$ iterations, indexed by $k=0,\ldots,K-1$. Each iteration trains one continuation candidate and $N_{\mathrm{ch}}$ design challengers. Subscripts identify the current state's components in~\eqref{eq:state}; superscripts $\mathrm{cont}$ and $\mathrm{ch}$ identify the two branches. Thus $K$ counts post-initialization iterations, not the initialization round.

\paragraph{Transfer boundary}
Within each task, morphology edits preserve kinematic connectivity, actuated-joint ordering, and the dimensions and meanings of observations and actions. Direct actor inheritance also requires the same policy-network architecture; matching vector dimensions alone does not establish compatibility. Geometry and training rewards may change within this interface. Topology changes and changes to observation or action semantics require a different transfer mechanism. Full-state continuation retains the morphology, reward, and learner configuration.

\begin{table}[!htb]
\caption{Training-state handling before each branch learns. Copied Actor parameters and observation statistics continue to update.}
\label{tab:state-transfer}\centering\footnotesize
\setlength{\tabcolsep}{3pt}
\begin{tabular}{@{}p{0.46\columnwidth}p{0.22\columnwidth}p{0.23\columnwidth}@{}}\hline
Component & Continuation & Challenger\\\hline
Morphology and reward & Unchanged & New proposal\\
Actor parameters & Restore & Copy\\
Observation statistics & Restore & Copy\\
Value / critic networks & Restore & Initialize anew\\
Optimizer states & Restore & Initialize anew\\
SAC targets and temperature & Restore & Initialize anew\\
SAC replay buffer & Restore & Empty\\
Environment / random generators & Restore & Initialize anew\\
Training counters & Continue & Reset\\\hline
\end{tabular}
\end{table}

\paragraph{Continuation Candidate}
The continuation candidate resumes $z_k$ on the current morphology $m_k$ and reward $r_k$ for another $B$ environment transitions:
\begin{align}
z_k^{\mathrm{cont}}&=\mathcal U_B^{r_k}(m_k;z_k),\\
y_k^{\mathrm{cont}}&=((m_k,z_k^{\mathrm{cont}}),r_k).
\label{eq:continuation}
\end{align}

\paragraph{Design Challengers}
CRAFT generates the ordered sequence of morphology--reward pairs
\begin{equation}
\mathcal Q_k=
\bigl((m_{k,j},r_{k,j})\bigr)_{j=1}^{N_{\mathrm{ch}}},
\label{eq:proposals}
\end{equation}
Here $m_{k,j}$ and $r_{k,j}$ specify challenger $j$, and $\mathcal Q_k$ is indexed in candidate order. Each challenger inherits the incumbent's actor and observation statistics within the transfer boundary above.

Let $\theta(z)$ denote the actor parameters. The operator $\operatorname{Init}(z_k)$ creates a template $\widetilde z_k$ with $\theta(\widetilde z_k)=\theta(z_k)$ and copied observation statistics. Both branches start from the incumbent state at the beginning of the iteration; challenger initialization does not use the continuation result. The copied normalizer preserves the input scaling expected by the inherited policy. Other learner components are initialized anew (Table~\ref{tab:state-transfer}), so value estimation and data collection adapt to the proposed dynamics and reward. Each challenger receives an independent copy of the template, and both its actor and normalizer continue to learn.

Starting from $\widetilde z_k$, the $j$-th design challenger performs reinforcement learning for $B$ environment transitions under morphology $m_{k,j}$ and reward $r_{k,j}$, yielding
\begin{align}
z_{k,j}^{\mathrm{ch}}
&=\mathcal U_B^{r_{k,j}}\left(m_{k,j};\widetilde z_k\right),\\
y_{k,j}^{\mathrm{ch}}&=((m_{k,j},z_{k,j}^{\mathrm{ch}}),r_{k,j}).
\label{eq:challenger}
\end{align}

\paragraph{Unified Evaluation and State Update}
Let $\mathcal C_k$ denote the comparison set of search states at iteration $k$:
\begin{equation}
\mathcal C_k=
\left\{
y_k,y_k^{\mathrm{cont}},y_{k,1}^{\mathrm{ch}},\ldots,
y_{k,N_{\mathrm{ch}}}^{\mathrm{ch}}
\right\}.
\label{eq:comparison}
\end{equation}
New candidates are evaluated using $(m,\Pi(z))$; the frozen incumbent keeps its recorded score. The next search state is
\begin{equation}
\begin{aligned}
\mathcal T_k&=\underset{y=((m,z),r)\in\mathcal C_k}{\arg\max}\;
\widehat S_\tau\bigl(m,\Pi(z)\bigr),\\
y_{k+1}&=\operatorname{first}_{\prec_k}(\mathcal T_k).
\end{aligned}
\label{eq:selection}
\end{equation}

Here $\mathcal T_k$ is the set of highest-scoring candidates, and $\operatorname{first}_{\prec_k}$ selects its first member under the fixed order $y_k\prec_k y_k^{\mathrm{cont}}\prec_k y_{k,1}^{\mathrm{ch}}\prec_k\cdots\prec_k y_{k,N_{\mathrm{ch}}}^{\mathrm{ch}}$. Thus a tied incumbent is retained; otherwise, the earliest highest-scoring new candidate is selected. Ties do not terminate the search. Incumbent retention makes the cached best score nondecreasing; it does not guarantee monotonic improvement in expected performance under fresh rollouts.
\subsection{CRAFT: Blackboard-Based Collaboration}

CRAFT coordinates four specialist LLM roles through a shared blackboard~\cite{hayes1985blackboard}. Given task $\tau$, the roles use prior evaluations and replayed behavior to propose morphology--reward pairs. The blackboard contains:
\begin{itemize}
  \item Experiment records $\mathcal A_k$. Each record $a_i=(y_i,s_i,b_i)$ identifies an evaluated state, its recorded score $s_i=\widehat S_\tau(m_i,\Pi(z_i))$, and its selection indicator $b_i$. Candidate and checkpoint identifiers link the result to the morphology, reward, and training files.
  \item Diagnostic evidence $\mathcal E_k$: the selected robot's replayed behavior and associated task measurements, used to diagnose motion and guide design revisions.
\end{itemize}
The shared input at iteration $k$ is
\begin{equation}
\mathcal B_k=(\mathcal A_k,\mathcal E_k).
\label{eq:blackboard}
\end{equation}

\begin{table}[!htb]
\caption{Shared CRAFT evidence and role-specific outputs.}
\label{tab:craft-shared-record}
\centering\footnotesize
\setlength{\tabcolsep}{3pt}
\renewcommand{\arraystretch}{1.08}
\begin{tabular}{@{}>{\raggedright\arraybackslash}p{0.27\columnwidth}>{\raggedright\arraybackslash}p{\dimexpr0.73\columnwidth-2\tabcolsep\relax}@{}}
\hline
\textbf{Record} & \textbf{Shared contents}\\\hline
Experiment history & Candidate identifier, selected score, and linked morphology, reward, and checkpoint files.\\
Current evidence & Saved replay video, trajectory and task measurements, and the selected-checkpoint reference.\\\hline
\textbf{Role} & \textbf{Output}\\\hline
Feedback & Failure hypotheses.\\
Morphology & Body edits and reward cross-review.\\
Reward & Reward code and body cross-review.\\
Integration & Paired morphology--reward candidates.\\\hline
\end{tabular}
\par\vspace{2pt}
\parbox{\columnwidth}{\raggedright\textit{Example history entry:} Ant, round 3; mainline continuation; selected score 60,870.41.}
\end{table}

Each role receives the same bounded view of $\mathcal A_k$, by default the latest five selection records, together with the current replay evidence and task measurements (Table~\ref{tab:craft-shared-record}). Within-round feedback and proposals pass directly between roles. A separate round-indexed archive $\mathcal H_k$ stores complete requests and responses; subsequent role inputs do not automatically include this archive. Diagnostic replay is distinct from the SAC replay buffer used for training.

For an initially trained robot $y_0=((m_0,z_0),r_0)$, its score and diagnostic rollout $\mathcal E_0$ establish
\begin{equation}
\mathcal B_0=\left(\{(y_0,\widehat S_\tau(m_0,\Pi(z_0)),1)\},\mathcal E_0\right).
\label{eq:blackboard-init}
\end{equation}
The indicator $1$ marks the selected initial robot. When no trained robot is available, an initialization round trains $N_{\mathrm{ch}}$ designs from scratch; their records and winner establish $(y_0,\mathcal B_0)$. Our benchmarks use six initial designs followed by four continuation--challenger rounds. The shared history remains fixed during each round and is updated after candidate evaluation.

\paragraph{Feedback and cross-review}
Feedback uses behavioral replays and task measurements to identify potential failure modes and propose coordinated morphology--reward changes. Available evidence comprises replay video frames and programmatically computed task measurements; Feedback produces the textual diagnoses passed to the design roles. The measurements include rollout length, cumulative reward, recorded root positions, and displacement. Morphology and Reward then draft independently:
\begin{align}
f_k&=\mathcal F(\tau,\mathcal B_k),\\
\mathcal M_k&=\mathcal G_M(\tau,\mathcal B_k,f_k,\varnothing),\\
\mathcal R_k&=\mathcal G_R(\tau,\mathcal B_k,f_k,\varnothing).
\end{align}
Here $\mathcal F,\mathcal G_M,\mathcal G_R$ are the Feedback, Morphology, and Reward calls; $f_k$ contains observed limitations and improvement hypotheses, while $\mathcal M_k$ and $\mathcal R_k$ are the initial design drafts. The final argument is review context; $\varnothing$ means no draft has yet been supplied. Both designers next receive both complete drafts, with their own first:
\begin{align}
\widehat{\mathcal M}_k&=\mathcal G_M(\tau,\mathcal B_k,f_k,(\mathcal M_k,\mathcal R_k)),\label{eq:morphology-review}\\
\widehat{\mathcal R}_k&=\mathcal G_R(\tau,\mathcal B_k,f_k,(\mathcal R_k,\mathcal M_k)).\label{eq:reward-review}
\end{align}
\begin{figure}[!t]
\small
\hrule\smallskip
\noindent\refstepcounter{lacealgorithm}\textbf{Algorithm~\thelacealgorithm}\quad LACE--CRAFT after initialization
\label{alg:lace-craft}
\smallskip\hrule\smallskip
{\raggedright\noindent\textbf{Input:} task $\tau$; initialized $y_0,\mathcal B_0,s(y_0)$;\\
\hspace*{1em}budget $B$; iterations $K$; challengers $N_{\rm ch}$.\par}
\begin{algorithmic}[1]
\STATE $y^*\leftarrow y_0$; $\mathcal H_0\leftarrow\varnothing$
\FOR{$k=0,\ldots,K-1$}
\STATE $((m_k,z_k),r_k)\leftarrow y_k$
\STATE $\widetilde z_k\leftarrow\operatorname{Init}(z_k)$ \COMMENT{Iteration-start Actor + normalizer}
\STATE $z_k^{\rm cont}\leftarrow\mathcal U_B^{r_k}(m_k;z_k)$
\STATE $y_k^{\rm cont}\leftarrow((m_k,z_k^{\rm cont}),r_k)$
\STATE $f_k\leftarrow\mathcal F(\tau,\mathcal B_k)$
\STATE $\mathcal M_k\leftarrow\mathcal G_M(\tau,\mathcal B_k,f_k,\varnothing)$
\STATE $\mathcal R_k\leftarrow\mathcal G_R(\tau,\mathcal B_k,f_k,\varnothing)$
\STATE $\widehat{\mathcal M}_k\leftarrow\mathcal G_M(\tau,\mathcal B_k,f_k,(\mathcal M_k,\mathcal R_k))$
\STATE $\widehat{\mathcal R}_k\leftarrow\mathcal G_R(\tau,\mathcal B_k,f_k,(\mathcal R_k,\mathcal M_k))$
\STATE $(\mathcal Q_k,h_k)\leftarrow\mathcal I(\tau,\mathcal B_k,f_k,\widehat{\mathcal M}_k,\widehat{\mathcal R}_k)$
\FOR{$j=1,\ldots,N_{\rm ch}$}
\STATE $(m_{k,j},r_{k,j})\leftarrow\mathcal Q_k[j]$
\STATE $z_{k,j}^{\rm ch}\leftarrow\mathcal U_B^{r_{k,j}}(m_{k,j};\widetilde z_k)$
\STATE $y_{k,j}^{\rm ch}\leftarrow((m_{k,j},z_{k,j}^{\rm ch}),r_{k,j})$
\ENDFOR
\STATE $\mathcal C_k\leftarrow\{y_k,y_k^{\rm cont}\}\cup\{y_{k,j}^{\rm ch}\}_{j=1}^{N_{\rm ch}}$
\FORALL{$y=((m,z),r)\in\mathcal C_k\setminus\{y_k\}$}
\STATE $s(y)\leftarrow\widehat S_\tau(m,\Pi(z))$
\ENDFOR
\STATE $\mathcal T_k\leftarrow\arg\max_{y\in\mathcal C_k}s(y)$
\STATE $y_{k+1}\leftarrow\operatorname{first}_{\prec_k}(\mathcal T_k)$; $y^*\leftarrow y_{k+1}$
\STATE $\Delta\mathcal A_k\leftarrow\{(y,s(y),\mathbf1[y=y_{k+1}]):y\in\mathcal C_k\}$
\STATE $\mathcal E_{k+1}\leftarrow\operatorname{Replay}(x_{k+1},\tau)$
\STATE $\mathcal B_{k+1}\leftarrow(\mathcal A_k\cup\Delta\mathcal A_k,\mathcal E_{k+1})$
\STATE $\mathcal H_{k+1}\leftarrow\mathcal H_k\cup\operatorname{Record}_H(k)$
\ENDFOR
\STATE \textbf{Return} $(m^*,\Pi(z^*))$, where $y^*=((m^*,z^*),r^*)$
\end{algorithmic}
\smallskip\hrule
\end{figure}

For the five-task benchmark, each reward proposal is complete Python/JAX source defining \texttt{compute\_reward}, following D2C~\cite{qiu2026debatecreate}. The callback maps observations, actions, and task metrics to a per-step training reward and named diagnostic terms; proposals may change both algebra and coefficients.

The hatted outputs are complete revised proposals. Through reciprocal critique~\cite{pmlr-v235-du24e,qiu2026debatecreate}, each designer assesses how the peer's changes affect its own proposal. These exchanges guide candidate generation; simulation supplies the evidence for selection.

\begin{figure*}[!t]
\centering
\includegraphics[width=0.98\textwidth]{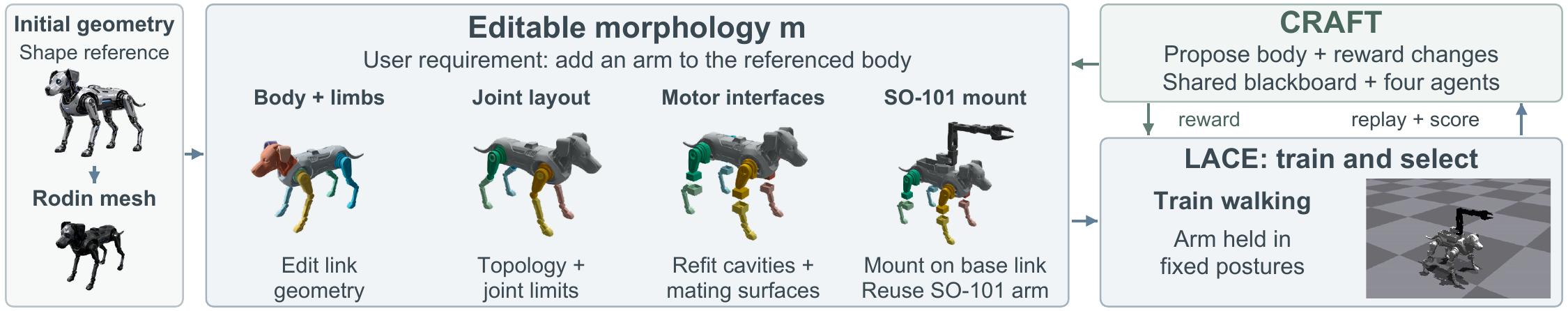}
\caption{An input image and user requirement define editable morphology $m$ through Rodin reconstruction and BANG decomposition. The four views show complementary components, not sequential steps. The base link incorporates an SO-101 mount, while other arm components reuse the SO-101 design. CRAFT supplies morphology edits and training rewards. LACE receives the compiled model and selects trained candidates using a task metric distinct from training reward. Replays and scores return to CRAFT. Simulation evaluates walking with fixed arm postures, not manipulation. Physical walking excludes the arm. The mounted-arm photograph documents assembly only.}
\label{fig:generative-morphology-flow}
\end{figure*}

\paragraph{Integration and records}
Integration combines the revisions into paired body--reward candidates:
\begin{equation}
(\mathcal Q_k,h_k)=\mathcal I
(\tau,\mathcal B_k,f_k,\widehat{\mathcal M}_k,\widehat{\mathcal R}_k).
\end{equation}
Here $\mathcal I$ is the Integration call, $\mathcal Q_k$ contains the $N_{\mathrm{ch}}$ pairs in~\eqref{eq:proposals}, and $h_k$ records their joint rationales. Integration can combine proposals with different source indices and revise either component. In the six-challenger configuration, two candidates receive each assessed design-risk label: Conservative, Medium, and Aggressive. All six follow the same inheritance rule and learning configuration.

\paragraph{Memory update}
After selection, $\Delta\mathcal A_k$ records the comparison states, scores, and selection flags. Deterministic policy inference on the selected robot produces the next diagnostic replay and task measurements:
\begin{equation}
\begin{aligned}
\mathcal E_{k+1}&=\operatorname{Replay}(x_{k+1},\tau),\\
\mathcal B_{k+1}&=(\mathcal A_k\cup\Delta\mathcal A_k,\mathcal E_{k+1}).
\end{aligned}
\label{eq:blackboard-update}
\end{equation}
The round's role requests and outputs are appended to $\mathcal H_k$ by $\operatorname{Record}_H(k)$, including feedback, drafts, revisions, and integration rationales. Candidate and round identifiers link records to the corresponding morphology, reward, and checkpoint. The next round uses the bounded experiment history and selected robot's replay evidence; the complete discussion archive is not automatically added to role inputs.

Algorithm~\ref{alg:lace-craft} summarizes the procedure. Its score map $s(y)$ stores $\widehat S_\tau(m,\Pi(z))$ for $y=((m,z),r)$; the incumbent's entry is reused. The indicator $\mathbf1[y=y_{k+1}]$ is one for the selected state and zero otherwise. The returned $y^*=((m^*,z^*),r^*)$ is the last retained state, or $y_0$ if no update occurs. Drafting and cross-review each use parallel role calls.

\subsection{Extension to Generative Robots}\label{sec:generative-extension}

The generative extension constructs an articulated morphology from a shape reference $c$ and user-specified hardware requirements, following the geometry-to-hardware approach of Text2Robot~\cite{ringel2025text2robot}. For an arm-equipped quadruped, the reference image guides body shape and the arm requirement specifies a torso mounting interface. Scale, orientation, and joint topology are configured alongside hardware $\eta$, which specifies motor dimensions, torque and speed limits, joint limits, control frequency, and payload. \bstctlcite{technical_urls_on}Hyper3D Rodin~\cite{hyper3d_docs} generates a static mesh from $c$; the model components in Fig.~\ref{fig:generative-morphology-flow} provide articulation and hardware interfaces. Task $\tau$ defines the evaluation objective.

\paragraph{Part decomposition and articulation}
BANG~\cite{zhang2025bang} decomposes the Rodin mesh into components. Scripts repair, merge, and scale them into body and limb meshes $\{g_i\}_{i=1}^{n}$, preserving part identifiers. Alignment scripts place the links in a common body frame, and an assembly template assigns them to a configured kinematic tree $G_m=(V,E)$~\cite{zhao2020robogrammar}. Each link stores its mesh, mass, center of mass, and inertia; each joint stores its type, parent-relative pose, axis, and limits under URDF conventions~\cite{urdf_kinetic_docs}. The illustrated geometry model has two actuated joints per leg. R17 denotes the hardware-adapted quadruped version used in the walking demonstration, with three actuated joints per leg.

\par\noindent\begin{minipage}{\columnwidth}
\paragraph{Actuator interfaces and export}
Scripts cut hip and knee motor cavities into aligned meshes using mesh Boolean operations and configured motor dimensions. Clearances separate mating surfaces; paired slot faces determine joint origins. The interface record $\Sigma$ stores motor poses, cavities, and mating surfaces for actuated joints. Screw holes and fastening details are designed manually. The torso (base link) incorporates an SO-101 mounting interface~\cite{Knight_Standard_Open_SO-100}\bstctlcite{technical_urls_off}; the arm links, motors, and gripper follow the existing SO-101 design. Export scripts construct the morphology package
\begin{equation}
\begin{split}
m^{\mathrm{gen}}&=(G_m,\Sigma,\Phi),\\
\Phi&=(\mathrm{URDF},\mathcal C_{\mathrm{coll}},\mathcal P_{\mathrm{inert}},\mathcal L,\Gamma).
\end{split}
\label{eq:generative-morphology}
\end{equation}
\end{minipage}\par
Here $\mathcal C_{\mathrm{coll}}$ and $\mathcal P_{\mathrm{inert}}$ contain collision and inertial assets computed from the processed geometry and configured material and motor masses. The fields $\mathcal L$ and $\Gamma$ specify joint and actuator limits, ordered action-to-joint mapping, action scaling, and control frequency. The training adapter loads this package into Isaac Gym for locomotion learning. In this backend, reward revisions update the task configuration's reward terms and parameters, which the PyTorch environment evaluates during training. For the arm-equipped robot, $\Gamma$ maps policy actions to the leg joints, while prescribed arm postures define the configurations under which walking is trained. Arm joints remain fixed during locomotion training.

\paragraph{Consistent morphology updates}
Morphology proposals can modify link dimensions and associated joint and mounting fields. Geometry scripts apply these edits, refit affected joint origins and motor slots, and regenerate collision and inertial assets, keeping the assembled robot model consistent with its simulation assets. Hardware $\eta$, kinematic connectivity, and the policy interface remain fixed, preserving compatibility with LACE's transfer rule. Fastening details require renewed manual checking after geometry changes.

\section{Experiments}
\begingroup
\setlength{\intextsep}{6pt plus 1pt minus 1pt}

\subsection{Experimental Setup}\label{sec:experimental-setup}

We evaluate on Ant, HalfCheetah, Hopper, Swimmer, and Walker2D using the D2C benchmark settings~\cite{freeman2021brax,qiu2026debatecreate}. Methods share each task's morphology and policy interfaces, learner, per-design training budget, horizon, and evaluation metric. Morphology search changes link geometry within a fixed kinematic tree. D2C is the primary co-design baseline; Default trains the original morphology and hand-designed reward from scratch.

D2C, LACE--CRAFT, and the ablation variants use GPT--5.6-sol and propose six morphology--reward pairs per round for five rounds, including initialization. Each search therefore generates 30 new designs. LACE additionally trains one continuation candidate in each of the four later rounds, giving 34 training units compared with D2C's 30.

Final performance is evaluated with seeds $0,1,2$, each using 128 parallel episodes, the task horizon, and the metric in~\eqref{eq:empirical-score}. We report the mean and sample standard deviation of the three seed-level mean scores. Morphology, policy parameters, and observation statistics remain frozen; actions are sampled from the learned policy. D2C returns a morphology--reward pair, from which a policy is trained with the task-specific budget; LACE--CRAFT returns a morphology--policy pair. These tests measure fixed-policy evaluation variability, not variation across training seeds or independent searches.

\subsection{Overall Performance against D2C}
\label{sec:main-results}
Figure~\ref{fig:d2c-final} normalizes the three-seed means by each task's D2C mean; Table~\ref{tab:raw-final} reports raw means and standard deviations. LACE--CRAFT has higher mean scores on all five tasks under the protocol above. Gains are 91.9\% on Ant, 36.8\% on HalfCheetah, 15.0\% on Hopper, 46.7\% on Swimmer, and 6.4\% on Walker2D.

\begin{figure}[H]
\centering
\includegraphics[width=0.90\columnwidth]{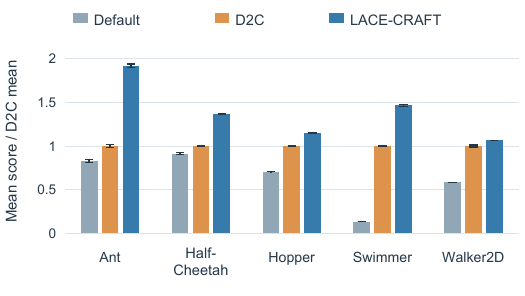}
\caption{Three-seed evaluation means with $\pm1$ sample-standard-deviation error bars. Means and deviations are divided by the task's fixed D2C mean; denominator uncertainty is not propagated.}
\label{fig:d2c-final}
\end{figure}

\begin{table}[!htb]
\caption{Final-policy evaluation scores: mean with $\pm$ sample standard deviation below, over three evaluation seeds (128 episodes each).}
\label{tab:raw-final}\centering\footnotesize
\begin{tabular}{@{}lrrr@{}}\hline
Task & Default & D2C & LACE--CRAFT\\\hline
Ant & \scorecell{27,362.80}{621.94} & \scorecell{33,044.51}{477.05} & \scorecell{63,396.82}{656.01}\\
HalfCheetah & \scorecell{18,343.94}{201.20} & \scorecell{20,072.15}{18.82} & \scorecell{27,455.51}{22.39}\\
Hopper & \scorecell{4,377.53}{9.80} & \scorecell{6,195.22}{2.29} & \scorecell{7,125.95}{11.94}\\
Swimmer & \scorecell{897.83}{8.19} & \scorecell{6,637.06}{10.67} & \scorecell{9,733.43}{68.55}\\
Walker2D & \scorecell{10,114.19}{10.37} & \scorecell{17,362.33}{221.69} & \scorecell{18,466.68}{29.89}\\\hline
\end{tabular}
\end{table}

\paragraph{A recorded search trajectory}
In the reported Ant search, round~2 selects a challenger scoring 54,646.39 over continuation at 40,341.80. Continuation then wins rounds~3 and~4 with 60,870.41 and 64,252.91, respectively. Round~5 retains the incumbent ahead of continuation at 63,058.92. These are historical selection scores, not the post-selection three-seed means. The trace illustrates how design adaptation, continued learning, and retention contribute within one search.

\subsection{Ablation Study}
\label{sec:ablations}
We examine policy reuse in LACE and behavioral feedback in CRAFT using two variants on all five tasks. \emph{No Actor inheritance} initializes challengers with fresh actor parameters and observation statistics. \emph{No replay evidence} excludes replay frames, trajectory-level evidence, and their derived behavioral diagnoses from all role inputs, while retaining scalar evaluation scores and design history. Both retain the four agents, shared blackboard, cross-review, and score history. Both ablations retain the full method's four continuation updates and per-branch training budget. The replay variant examines behavioral context within CRAFT.

\begin{figure}[H]
\centering
\includegraphics[width=0.90\columnwidth]{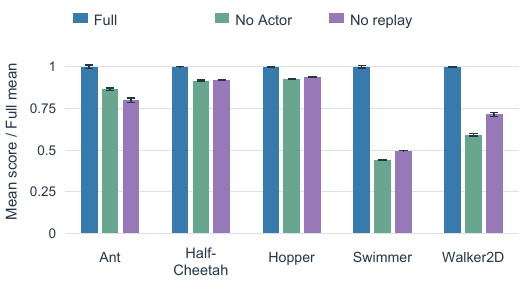}
\caption{Three-seed evaluation means with $\pm1$ sample-standard-deviation error bars, divided by the task's fixed full-method mean. Denominator uncertainty is not propagated.}
\label{fig:ablations}
\end{figure}

\begin{table}[H]
\centering\footnotesize
\caption{Final-policy ablations: mean with $\pm$ sample standard deviation below, over three evaluation seeds (128 episodes each).}
\label{tab:ablation-scores}
\setlength{\tabcolsep}{4pt}
\begin{tabular}{@{}lrrr@{}}\hline
Task & Full & No Actor & No replay\\\hline
Ant & \scorecell{63,396.82}{656.01} & \scorecell{54,806.01}{432.08} & \scorecell{50,609.18}{838.12}\\
HalfCheetah & \scorecell{27,455.51}{22.39} & \scorecell{25,147.09}{116.61} & \scorecell{25,272.34}{25.86}\\
Hopper & \scorecell{7,125.95}{11.94} & \scorecell{6,600.18}{9.61} & \scorecell{6,693.74}{2.87}\\
Swimmer & \scorecell{9,733.43}{68.55} & \scorecell{4,278.96}{23.92} & \scorecell{4,832.12}{2.60}\\
Walker2D & \scorecell{18,466.68}{29.89} & \scorecell{10,919.33}{133.34} & \scorecell{13,171.87}{232.93}\\\hline
\end{tabular}
\end{table}

\paragraph{Policy reuse in LACE}
The full method has a higher three-seed mean than No Actor on all five tasks (Fig.~\ref{fig:ablations}). The largest gaps occur on Swimmer and Walker2D, where No Actor reaches 44.0\% and 59.1\% of the full-method mean, respectively. The full method's advantage is smaller on Ant, HalfCheetah, and Hopper, at 8.0--15.7\%. These comparisons concern the transferred actor and observation statistics together.

\paragraph{Behavioral feedback in CRAFT}
The full method also has a higher three-seed mean than No replay on all five tasks. The relative gap is largest on Swimmer, where No replay reaches 49.6\% of the full-method mean, and smallest on Hopper, at 93.9\%. The no-replay condition remains closer to the full method on Hopper than on Swimmer, indicating a task-dependent difference for these selected policies.

\subsection{Generative Robot Demonstration}

\begin{figure}[!htb]
\centering
\includegraphics[width=0.98\columnwidth]{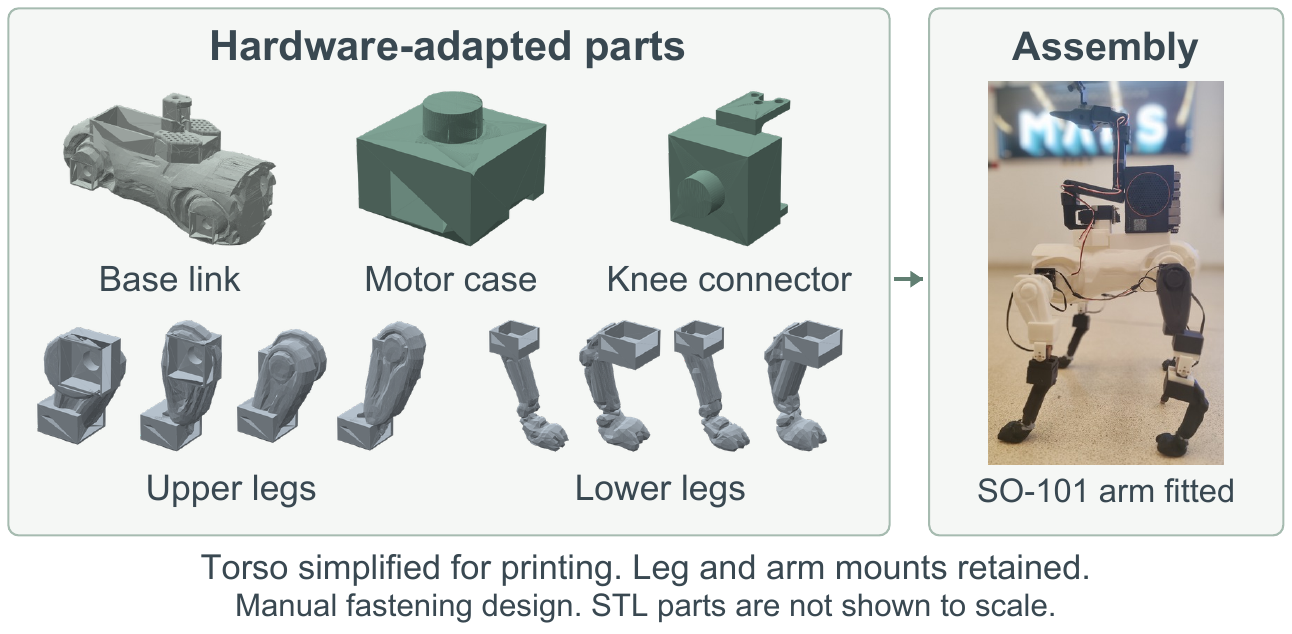}
\caption{Hardware-adapted printable parts and assembly with the SO-101 arm.}
\label{fig:hardware-fabrication}
\end{figure}

The fabricated quadruped demonstrates the conversion of generated geometry into printable parts and a robot assembled with physical actuators (Fig.~\ref{fig:hardware-fabrication}). Its base link retains the leg attachments and an SO-101 arm mount, while the head and tail are omitted to simplify printing and reduce cost. The printed torso therefore differs from the simulated geometry.

The mounted-arm photograph documents the hardware interface; the walking trial uses the arm-free configuration. An onboard Raspberry Pi runs the R17 PPO controller at 20 Hz. The controller was trained in the Isaac Gym hardware-adaptation task and maps 12 policy outputs to the leg-joint position commands.
 Figure~\ref{fig:assembled-generative-robot} shows consecutive stages of indoor walking. This provides qualitative evidence of fabrication and locomotion, while quantitative transfer from the co-design search and walking with the arm remain to be evaluated.

\begin{figure}[H]
\centering
\includegraphics[width=0.70\columnwidth]{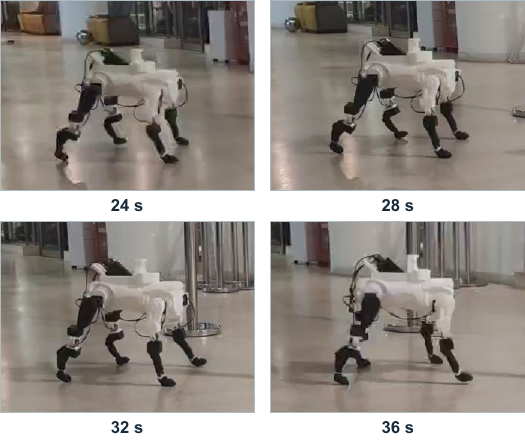}
\caption{Indoor walking with the arm removed. Frames are from one video; timestamps are in seconds.}
\label{fig:assembled-generative-robot}
\end{figure}

\endgroup

\FloatBarrier
\section{Limitations and Future Work}\label{sec:limitations}

\paragraph{Interpreting replay evidence}\label{sec:replay-limitations}
Behavioral replays can support multiple explanations, so task measurements should corroborate proposed diagnoses. The three evaluation seeds quantify rollout variability of fixed policies; they do not quantify variation across complete searches or training runs, or disentangle errors in evidence interpretation from variation in proposal generation.

\paragraph{Fabrication and hardware integration}
Fabricating the platform requires manual screw-hole and fastening design (Fig.~\ref{fig:hardware-fabrication}). The printed parts underwent no structural simulation, and thin connections were prone to fracture. Structural analysis under expected loads and durability testing are needed to improve mechanical reliability.

\paragraph{Closing the sim-to-real loop}
Differences between the simulated model and fabricated robot remain a challenge for policy transfer. Incorporating hardware video and measured joint motions into CRAFT's shared blackboard would allow subsequent design revisions to draw on both simulation and physical execution.

\FloatBarrier
\section{Conclusion}
LACE--CRAFT combines policy reuse with morphology--reward search guided by behavioral evidence. LACE compares adapted challengers, full-state continuation, and incumbent retention under a fixed task metric. CRAFT coordinates proposals and cross-review through shared experimental records and replays. The selected LACE--CRAFT policies achieve higher three-seed evaluation means than D2C on all five tasks under the stated protocol. The generative extension connects editable geometry to articulated models and simulation assets; the fabricated prototype demonstrates indoor walking with the arm removed. Future work will address diagnostic reliability, mechanical validation, and quantitative sim-to-real evaluation.

\section*{Acknowledgments}
This work was supported in part by the National Natural Science Foundation of China under Grant W2431046, the National Key R\&D Program of China under Grant 2025YFA1309603, the Central Guided Local Science and Technology Foundation of China under Grant YDZX20253100001001, the MoE Key Lab of Intelligent Perception and Human-Machine Collaboration (ShanghaiTech University), and the Shanghai Frontiers Science Center of Human-centered Artificial Intelligence.

\bibliographystyle{IEEEtran}
\bibliography{scholar_export_received,bibliography_layout,technical_references_20260915}
\clearpage
\useRomanappendicesfalse
\appendices
\numberwithin{equation}{section}
\setcounter{table}{0}
\renewcommand{\thetable}{S\arabic{table}}
\setcounter{figure}{0}
\renewcommand{\thefigure}{S\arabic{figure}}
\setlength{\intextsep}{8pt plus 2pt minus 2pt}

\section{Training-State Transfer}\label{app:state-transfer}

\subsection{Continuation, inheritance, and incumbent retention}
LACE carries three distinct states into candidate comparison. The incumbent retains its morphology, reward, checkpoint, and recorded evaluation score without further training. Continuation restores the complete learner on that morphology and reward. A design challenger initializes a new learner from the incumbent's Actor and observation statistics, then trains on the proposed morphology--reward pair. Table~\ref{tab:supp-state-transfer} lists the component-level operations.

\begin{table}[H]
\centering\footnotesize
\caption{Checkpoint handling for the current LACE protocol.}
\label{tab:supp-state-transfer}
\setlength{\tabcolsep}{3pt}
\begin{tabular}{@{}p{0.43\columnwidth}p{0.23\columnwidth}p{0.25\columnwidth}@{}}\hline
Component & Continuation & Challenger\\\hline
Morphology and reward & Unchanged & Proposed pair\\[2pt]
Actor parameters & Restore & Copy\\[2pt]
Observation statistics & Restore & Copy\\[2pt]
Value / critic parameters & Restore & Initialize anew\\[2pt]
Optimizer states & Restore & Initialize anew\\[2pt]
SAC targets and entropy state & Restore & Initialize anew\\[2pt]
SAC replay buffer & Restore & Empty buffer\\[2pt]
Environment and random generators & Restore & Initialize anew\\[2pt]
Training counters & Continue & New training unit\\\hline
\end{tabular}
\end{table}

Copied Actor parameters and observation statistics continue to update during challenger training. Each challenger receives its own learner state, so training one candidate does not modify the initialization used by another. A copied normalizer preserves the observation scaling expected by the inherited policy at the start of the new training unit.

The frozen incumbent remains in the comparison set with its cached score. Ties are resolved in the fixed order of incumbent, continuation, and ordered challengers; a tied incumbent is retained, and ties do not terminate the search. This rule makes retained evaluation scores nondecreasing along the search, while fresh evaluations of a retained policy can still vary.

\subsection{PPO continuation state}
The serialized PPO state is represented by
\begin{equation}
z^{\mathrm{PPO}}=(\theta_\pi,\theta_V,\nu,o,c_e,\xi).
\label{eq:supp-ppo-state}
\end{equation}
Here $\theta_\pi$ and $\theta_V$ are Actor and value-network parameters, $\nu$ contains optimizer variables, $o$ contains observation-normalization statistics, and $c_e$ counts environment transitions. The collection state $\xi$ contains the wrapped environment state and the random-generator states used by learning, rollout collection, and evaluation. The continuation implementation serializes these variables around the Brax PPO learner~\cite{freeman2021brax,schulman2017proximal}.

Continuation restores all components of~\eqref{eq:supp-ppo-state}. Challenger initialization copies $\theta_\pi$ and $o$, creates a new value function and optimizer, resets collection state and counters, and collects trajectories on the new body. Value estimates are therefore learned under the challenger's dynamics and training reward.

\subsection{SAC continuation state}
For SAC~\cite{haarnoja2018soft}, the serialized state is
\begin{equation}
\begin{split}
z^{\mathrm{SAC}}=(&\theta_\pi,\theta_Q,\bar\theta_Q,\alpha,
\nu_\pi,\nu_Q,\nu_\alpha,\\
&o,c_e,c_g,\mathcal D,\xi).
\end{split}
\label{eq:supp-sac-state}
\end{equation}
The critics and target critics have parameters $\theta_Q$ and $\bar\theta_Q$; $\alpha$ is the entropy-temperature state. The variables $\nu_\pi$, $\nu_Q$, and $\nu_\alpha$ are the corresponding optimizer states. The counter $c_g$ records gradient updates, and $\mathcal D$ is the replay-buffer state. The remaining variables have the meanings given for PPO.

SAC continuation restores the critics, targets, entropy state, and collected replay data together with the Actor. A challenger initializes these components anew and starts with an empty replay buffer. Thus its critic updates use transitions collected under its own morphology and reward.

\subsection{Policy-interface compatibility}
The benchmark morphology adapters preserve observation dimensions, action dimensions, and actuated-joint ordering within each task. Inheritance uses a direct parameter copy between compatible policy networks. All six challengers in the current protocol receive the same Actor and normalizer transfer; Conservative, Medium, and Aggressive denote design-risk groups, not transfer coefficients.

When initialization starts from six untrained designs, all six use fresh learners. Actor inheritance begins after selection of the first trained robot. Task-specific initialization settings are detailed in Appendix~\ref{app:eval-budget}.

\section{CRAFT Requests, Responses, and Persistent Records}\label{app:craft-records}

\subsection{Role requests and execution order}
Every role receives the task description, editable morphology interface, training configuration, and a bounded blackboard view. The standard view contains the latest five selection records and the selected robot's current numerical behavior summary and artifact references. This shared input remains fixed during the round.

Feedback executes first. Morphology and Reward then draft in parallel, each receiving the Feedback response. Once both drafts are available, the two designers cross-review in parallel. Each review receives both complete drafts, with the receiving role's draft first. Integration executes after both revisions are available.

Let $p_k^a$ be the input packet for role $a$ in round $k$, and let $\iota_a$ contain its role-specific instructions. With superscripts $0$ and $\mathrm{rev}$ identifying initial drafting and cross-review, the packets are
\begin{align}
p_k^F&=(\tau,\mathcal B_k,\iota_F),\\
p_k^{M,0}&=(\tau,\mathcal B_k,f_k,\iota_M),\\
p_k^{R,0}&=(\tau,\mathcal B_k,f_k,\iota_R),\\
p_k^{M,\mathrm{rev}}&=(\tau,\mathcal B_k,f_k,(\mathcal M_k,\mathcal R_k),
\iota_M^{\mathrm{rev}}),\\
p_k^{R,\mathrm{rev}}&=(\tau,\mathcal B_k,f_k,(\mathcal R_k,\mathcal M_k),
\iota_R^{\mathrm{rev}}),\\
p_k^I&=(\tau,\mathcal B_k,f_k,\widehat{\mathcal M}_k,
\widehat{\mathcal R}_k,\iota_I).
\label{eq:supp-role-packets}
\end{align}
The quantities $f_k$, $\mathcal M_k$, $\mathcal R_k$, and their hatted revisions are the outputs defined in the method. A round without retries uses six role calls: one Feedback call, two drafts, two reviews, and one Integration call.

\subsection{Output contracts}
\paragraph{Feedback}
The response identifies limitations in the recorded behavior and proposes testable design changes. Its evidence combines replay video frames with programmatically computed task measurements, including rollout length, cumulative reward, root-position summaries, displacement, and numerical-finiteness checks. Explanations involving balance, contact, or failure mechanisms remain hypotheses when those quantities are not directly measured.

\paragraph{Morphology}
The draft contains at least six complete morphology vectors within the task's editable interface. Each vector is paired with a rationale connecting its geometric changes to the task and Feedback response. The revised output contains complete proposals after considering the Reward draft.

\paragraph{Reward}
The draft contains at least six complete Python/JAX programs. Each JSON proposal supplies \texttt{proposal\_id}, \texttt{reward\_program}, \texttt{mechanism}, and \texttt{scale\_audit}. The code defines the D2C reward callback~\cite{qiu2026debatecreate}:
\begin{quote}\footnotesize\ttfamily
compute\_reward(obs, action, prev\_action,\\
\hspace*{1em}dt, metrics) -> (reward, extras)
\end{quote}
Here \texttt{obs} is the environment observation supplied at that transition, \texttt{action} and \texttt{prev\_action} are the current and previous controls, \texttt{dt} is the control-step duration, and \texttt{metrics} contains the task-specific quantities exposed by the environment. The output \texttt{reward} is the per-step scalar optimized by RL; \texttt{extras} maps component names to diagnostic values. Available metric keys are task-dependent. For Ant they include \texttt{forward\_reward}, \texttt{healthy\_reward}, \texttt{ctrl\_cost}, \texttt{contact\_cost}, and \texttt{is\_healthy}.

The proposal specifies the full expression, including coefficients, nonlinearities, and couplings. For instance, one recorded HalfCheetah candidate computes \texttt{metrics['x\_velocity']} minus $0.014$ times the squared action norm, returning the velocity and negative effort term separately for logging. This is a candidate-specific expression, not a reward template imposed on all tasks. Cross-review revises the complete programs in response to morphology drafts; Integration emits the final program paired with each geometry vector.

The runtime parses the source and checks task-specific metric access before saving \texttt{reward.py} alongside \texttt{morphology.xml}. Their hashes are stored together in the candidate binding. The D2C compiler loads the callable into the environment. Its compatibility adapter accepts scalar-only returns or reduced signatures and normalizes them to a reward--diagnostics pair; this does not change the full-source proposal format. The external selection metric is computed separately from the generated reward.

\paragraph{Integration}
Integration returns exactly six complete morphology--reward pairs, with two assigned to each design-risk group. It can combine proposals with different source indices and revise either component. Each result contains a unique candidate identifier, source proposal identifiers, the complete morphology and reward, a joint rationale, and a risk assessment.

The structured output is checked for candidate count, unique identifiers, required risk groups, renderable geometry, and the reward-program interface. A rejected response is stored with its validation error, and the next request asks for a complete corrected response. The role pipeline permits up to three stage-level attempts. Accepted candidates subsequently undergo training and external evaluation.

\subsection{Blackboard and archive contents}
The blackboard $\mathcal B_k=(\mathcal A_k,\mathcal E_k)$ combines stored experiment records with the selected checkpoint's diagnostic evidence. Candidate and checkpoint identifiers connect scores to morphology files, reward source, and training state. Large replay artifacts remain in files; role requests contain their references and the numerical summary.

\begin{table}[H]
\centering\footnotesize
\vspace*{7pt}
\caption{Persistent records and their use in CRAFT.}
\label{tab:supp-record-fields}
\begin{tabular}{@{}p{0.25\columnwidth}p{0.68\columnwidth}@{}}\hline
Record & Stored content and use\\\hline
Selected robot & Candidate and checkpoint identifiers, morphology and reward bindings, score, and full-state checkpoint.\\[3pt]
Selection history $\mathcal A_k$ & Compared candidates, scores, and round-specific selection flags; the bounded view enters the next discussion.\\[3pt]
Evidence $\mathcal E_k$ & Selected-checkpoint rollout summary and references to replay and visual manifests.\\[3pt]
Role archive $\mathcal H_k$ & Requests, responses, drafts, revisions, integration rationales, validation errors, and API records; retained separately from future role inputs.\\\hline
\end{tabular}
\end{table}

New Feedback and proposal text is passed directly between roles within a round. After training and selection, the comparison results are appended to $\mathcal A_k$, and a diagnostic rollout produces $\mathcal E_{k+1}$ for the selected robot. The separate archive $\mathcal H_k$ preserves the complete discussion without automatically expanding the next round's language-model context.

\subsection{Linking results, proposals, and replay}
Each selection record associates a round and candidate identifier with an evaluation score, a selection flag, and a checkpoint. The candidate's asset record identifies its morphology XML and reward program. This association distinguishes a new body--reward proposal from further training of an existing design.

Replay evidence is indexed by the selected checkpoint. Its manifest links the numerical rollout summary, saved HTML replay, and morphology assets; a file hash identifies the corresponding replay. Drafts, cross-reviews, and integration responses are indexed by round and role in $\mathcal H_k$. These links connect the measured behavior to the robot that produced it and the proposals generated in that discussion.

\section{Evaluation and Training Budgets}\label{app:eval-budget}

\subsection{Selection metric and diagnostic measurements}
Candidate selection uses \texttt{eval/episode\_distance\_from\_origin}: the root-position norm in the task coordinates listed in Table~\ref{tab:supp-five-init}. Values are summed over the episode and averaged across evaluation episodes. This metric measures accumulated distance from the origin, rather than path length, signed forward displacement, or training return.

The diagnostic replay is generated separately. It loads the selected morphology, reward, and policy parameters and runs deterministic inference with a reset seed derived from the checkpoint identifier. The rollout ends at termination or the task horizon. Pipeline-state frames, including the initial state, are saved for the HTML replay.

The numerical replay summary records executed steps, cumulative reward, the first and last recorded root $x$ coordinates, their absolute difference, and a finiteness flag. These positions are sampled after environment steps; the initial rendered frame is stored separately. The coordinate difference is a displacement diagnostic. Numerical finiteness checks valid numerical values and does not establish balance or task success. Replay video frames and these numerical measurements are available as behavioral evidence; Feedback produces the textual diagnoses passed to the design roles.

\subsection{Learner settings}
Tables~\ref{tab:supp-ppo-config} and~\ref{tab:supp-sac-config} give algorithm-specific settings. Observation normalization is enabled, action repeat is one, and each task uses 128 parallel evaluation environments. The seeds are RL seeds for the individual searches.

\begin{table}[H]
\centering\footnotesize
\caption{PPO configuration. HC denotes HalfCheetah.}
\label{tab:supp-ppo-config}
\setlength{\tabcolsep}{3pt}
\begin{tabular}{@{}lrrr@{}}\hline
Setting & Ant & HC & Swimmer\\\hline
Learning rate & $3\times10^{-4}$ & $3\times10^{-4}$ & $3\times10^{-4}$\\
Discount & 0.97 & 0.95 & 0.97\\
Entropy cost & 0.01 & 0.001 & 0.001\\
Reward scaling & 25 & 3 & 5\\
Training environments & 256 & 128 & 128\\
Batch size & 512 & 512 & 128\\
Unroll length & 5 & 15 & 6\\
Minibatches & 8 & 16 & 8\\
Updates per batch & 4 & 8 & 4\\
Episode horizon & 450 & 400 & 600\\
RL seed & 0 & 0 & 0\\\hline
\end{tabular}
\end{table}

\begin{table}[H]
\centering\footnotesize
\caption{SAC configuration.}
\label{tab:supp-sac-config}
\setlength{\tabcolsep}{4pt}
\begin{tabular}{@{}lrr@{}}\hline
Setting & Hopper & Walker2D\\\hline
Learning rate & $6\times10^{-4}$ & $6\times10^{-4}$\\
Discount & 0.997 & 0.997\\
Reward scaling & 30 & 5\\
Training environments & 12 & 12\\
Batch size & 48 & 48\\
Gradient updates per step & 3 & 3\\
Minimum replay size & 2,048 & 4,096\\
Maximum replay size & 50,000 & 65,536\\
Episode horizon & 800 & 1,000\\
RL seed & 1 & 1\\\hline
\end{tabular}
\end{table}

Requested transition budgets are aligned to complete learner batches (Table~\ref{tab:task-settings}). Training units use these endpoints rather than truncating the last learner batch.

\begin{table}[H]
\vspace*{7pt}
\caption{Task-specific learning configurations. Actual transition counts include batch alignment.}
\label{tab:task-settings}\centering\footnotesize
\setlength{\tabcolsep}{2.4pt}
\begin{tabular}{lcrrr}\hline
Task & RL & Requested & Actual & Horizon\\\hline
Ant & PPO & 8,000,000 & 8,007,680 & 450\\
HalfCheetah & PPO & 5,000,000 & 5,038,080 & 400\\
Hopper & SAC & 400,000 & 400,008 & 800\\
Swimmer & PPO & 350,000 & 350,208 & 600\\
Walker2D & SAC & 500,000 & 500,004 & 1,000\\\hline
\end{tabular}
\end{table}

\subsection{Networks and software}
PPO uses an actor with four 32-unit hidden layers and a value network with five 256-unit layers, both with Swish activation. SAC uses two 256-unit ReLU layers for the actor and each of two critics. Actors parameterize a tanh-squashed Gaussian. The five-task benchmark uses Python 3.10.20, Brax 0.12.4, JAX/JAXlib 0.6.2, Flax 0.10.6, MuJoCo/MJX 3.3.3, and Optax 0.2.8. This software configuration applies to the five-task benchmark; generated robots use the Isaac Gym adapter described in Sec.~\ref{sec:generative-extension}.

\subsection{Default baseline and reported checkpoint scores}
Default uses the task's original morphology and built-in hand-designed reward, with morphology and reward generation disabled and no LLM calls. Each policy is initialized from scratch and receives one task-specific training budget: 8 million transitions for Ant, 5 million for HalfCheetah, 400,000 for Hopper, 350,000 for Swimmer, and 500,000 for Walker2D. The PPO/SAC hyperparameters and task horizons follow the D2C-aligned settings above; Default is not a 30-candidate search.

The reported Default checkpoints use seeds 0, 0, 1, 0, and 2 for Ant, HalfCheetah, Hopper, Swimmer, and Walker2D, respectively. The historical Walker2D value $10{,}125.74$ is the stored seed-2 checkpoint score, not the three-seed evaluation mean reported in the main text. Its implementation uses NaN-safe masks for inactive evaluation episodes and rejected MJX line-search updates. Morphology, reward, and transition budget remain unchanged; these numerical repairs are part of the implementation configuration.

Training and evaluation use separate simulator rollouts. At the training endpoint, the evaluator holds the policy fixed and runs 128 parallel episodes with independently reset environments, the configured task horizon, and the task termination rule. Policy sampling follows the learner's evaluation setting; the PPO implementation uses stochastic evaluation by default. The evaluator logs the episode mean of the fixed task metric as \texttt{eval/episode\_distance\_from\_origin}. This value is stored in the candidate result and used for selection. A retained incumbent keeps its recorded score. The historical search traces in this appendix report these stored selection scores. The main comparison and ablation tables instead report post-selection evaluation: the mean and sample standard deviation across evaluation seeds $0,1,2$, each with 128 parallel episodes and frozen policy parameters, morphology, and observation statistics. These evaluations do not retrain the selected LACE--CRAFT policies.

These evaluation episodes estimate the performance of one trained policy. Repeating the complete proposal--training--selection search measures a different source of variation. D2C reports multiple search seeds and separate retraining studies~\cite{qiu2026debatecreate}; adopting its per-candidate learner settings does not imply those repetitions in the present main tables. The Ant search-level analysis below reports the complete searches available for that comparison.

\subsection{Morphology parameter bounds}
Morphology vectors contain 10, 24, 10, 6, and 10 scalars for Ant, HalfCheetah, Hopper, Swimmer, and Walker2D, respectively. All coordinates and radii below are in meters. The indexed scalars $p_i$ are the entries of the task's morphology vector, not policy parameters. Bounds are applied before rendering; additional geometric constraints reject combinations that are individually in range but do not form valid bodies.

\paragraph{Ant}
The torso radius is $p_1\in[.05,.50]$. Proximal planar offsets $p_{2:5}$ each lie in $[.05,.80]$, distal offsets $p_{6:7}$ in $[.05,1.50]$, and capsule radii $p_{8:10}$ in $[.01,.30]$. Mirroring these offsets constructs the four limbs. Each link radius is at most half its segment length, with nominal ground clearance checked at the configured reset pose.

\paragraph{HalfCheetah}
Coordinates $p_{1:4}$ describe the torso endpoints and head endpoint. Rear-leg endpoint offsets are $p_{5:10}$, and front-leg offsets are $p_{11:16}$, ordered as thigh, shin, and foot $(x,z)$ pairs. Radii $p_{17:24}$ correspond to torso, head, rear thigh/shin/foot, and front thigh/shin/foot. Table~\ref{tab:supp-hc-bounds} gives every interval. Joint ranges and actuator gears remain fixed, and initialization height is derived to preserve ground clearance.
\begin{table}[H]
\centering\footnotesize
\caption{HalfCheetah morphology-vector intervals.}
\label{tab:supp-hc-bounds}
\setlength{\tabcolsep}{7pt}
\begin{tabular}{@{}clcl@{}}\hline
Index & Interval & Index & Interval\\\hline
1 & $[-.90,-.20]$ & 13 & $[-.35,.50]$\\
2 & $[.20,.90]$ & 14 & $[-.45,-.04]$\\
3 & $[.55,1.35]$ & 15 & $[-.50,.35]$\\
4 & $[.04,.45]$ & 16 & $[-.35,-.02]$\\
5 & $[-.50,.30]$ & 17 & $[.015,.100]$\\
6 & $[-.45,-.04]$ & 18 & $[.010,.090]$\\
7 & $[-.50,.35]$ & 19 & $[.010,.090]$\\
8 & $[-.45,-.04]$ & 20 & $[.010,.090]$\\
9 & $[-.35,.50]$ & 21 & $[.008,.080]$\\
10 & $[-.35,-.02]$ & 22 & $[.010,.090]$\\
11 & $[-.30,.50]$ & 23 & $[.010,.090]$\\
12 & $[-.45,-.04]$ & 24 & $[.008,.080]$\\\hline
\end{tabular}
\end{table}

\paragraph{Hopper}
Vertical anchors $p_{1:4}$ have intervals $[.3,3]$, $[.2,2.5]$, $[.1,2]$, and $[.01,1.5]$, with $p_1>p_2>p_3>p_4>0$. Foot endpoints satisfy $p_5\in[.01,.5]$ and $p_6\in[-.8,-.01]$. The four radii $p_{7:10}$ lie in $[.01,.2]$. Each segment length must exceed its diameter, and the canonical pose must clear the floor.

\paragraph{Swimmer}
The three segment lengths $p_{1:3}$ each lie in $[.1,3]$, and their radii $p_{4:6}$ in $[.01,.5]$. The generator retains the ordered three-link chain and two actuated joints.

\paragraph{Walker2D}
Vertical anchors $p_{1:4}$ have intervals $[.3,3]$, $[.2,2.5]$, $[.1,2]$, and $[.01,1.5]$, with $p_1>p_2>p_3>p_4>0$. Here $p_1$ is the torso center. Foot endpoints satisfy $p_5\in[.01,.5]$ and $p_6\in[-.8,0]$; radii $p_{7:10}$ lie in $[.01,.2]$. Additional checks impose $0.8<p_1<2$, segment lengths of at least $0.1$, radii no greater than $0.45$ times the corresponding segment length, and nonnegative foot clearance.

\subsection{Model calls and failed candidates}
The recorded model identifier is GPT-5.6-sol with temperature 1.0, low reasoning effort, and low verbosity. The client requests a JSON object through a streaming Responses-compatible interface, with a Chat-Completions-compatible fallback. No explicit maximum output-token count is set. The CRAFT role pipeline allows three attempts per role; the transport client also has retry handling. A six-role-call round is therefore a logical call count, not a limit on metered API requests.

Validation checks the morphology vector, generated XML, reward syntax and interface, candidate identifiers, and six-candidate portfolio. Invalid outputs retain their error records for correction; exhaustion halts the round. A training exception or non-finite evaluation score leaves the round unselected. The pipeline does not assign a synthetic zero score to a failed candidate. Resumption preserves already completed, checkpoint-bound results. Training budgets count completed units in the reported search.

\subsection{Design counts and total training}
D2C and LACE--CRAFT generate six new designs per round for five rounds, giving 30 designs per search. LACE trains six initial designs, then six challengers and one continuation in each of four later rounds:
\begin{equation}
N_{\mathrm{train}}=6+4(6+1)=34.
\label{eq:supp-training-count}
\end{equation}
Here $N_{\mathrm{train}}$ counts completed training units. Incumbent retention uses a cached score and adds no training unit. The D2C comparison matches new-design counts and per-unit settings, with LACE using four additional continuation units. The original D2C study used an 80-policy, multi-seed protocol~\cite{qiu2026debatecreate}; the results here use the stated 30-design reproduction.

\subsection{Five-task initialization configurations}
Table~\ref{tab:supp-five-init} records the initialization protocol for all five tasks. Each search begins with six fresh designs; subsequent rounds copy the incumbent Actor parameters and observation statistics into compatible challengers and restore the complete saved learner state for continuation. Conservative, Medium, and Aggressive labels describe proposal risk, not different transfer coefficients.
\begin{table}[H]
\centering\footnotesize
\caption{Initialization of the reported five-task searches.}
\label{tab:supp-five-init}
\setlength{\tabcolsep}{3pt}
\begin{tabular}{@{}lcp{.32\columnwidth}p{.22\columnwidth}@{}}\hline
Task & Coordinates & Challenger & Continuation\\\hline
Ant & $(x,y,z)$ & Copy & Full state\\
HalfCheetah & $(x,y,z)$ & Copy & Full state\\
Hopper & $(x,y,z)$ & Copy & Full state\\
Swimmer & $(x,y)$ & Copy & Full state\\
Walker2D & $(x,y,z)$ & Copy & Full state\\\hline
\end{tabular}
\end{table}

\subsection{Search-level variability and normalized visualization}
The main five-task table reports three-seed evaluation means and sample standard deviations for the final policies. Its normalized plot divides means and deviations by the fixed D2C mean within each task; the ablation plot uses the fixed full-method mean. Denominator uncertainty is not propagated. Neither normalization pools episodes or tasks into statistical replicates. The historical search-selection scores below are a separate record.

The Ant variability analysis uses three complete searches with the same task configuration, role-input specification, PPO settings, and inheritance rules. Each independently generates six initial designs and completes four continuation--challenger rounds. Their final scores are 57,560.79, 55,259.83, and 64,252.91; the latest search supplies the selected Full checkpoint subsequently evaluated for the main comparison tables. For $n=3$ searches with round-end scores $s_{i,k}$, the plotted center and spread are $\bar s_k=n^{-1}\sum_i s_{i,k}$ and $\sqrt{\sum_i(s_{i,k}-\bar s_k)^2/(n-1)}$. The final mean and sample standard deviation are $59{,}024.51\pm4{,}671.80$, not confidence limits. The RL seed is fixed at zero, so this comparison measures variation between complete searches under that training seed.

Each Ant training unit contains 8,007,680 actual transitions. LACE's plotted budgets are therefore 48,046,080; 104,099,840; 160,153,600; 216,207,360; and 272,261,120 transitions. D2C uses 48,046,080; 96,092,160; 144,138,240; 192,184,320; and 240,230,400. Points mark completed rounds; connecting segments do not represent additional evaluations. Cached incumbent scores add no training budget. The figure does not count generation latency, inference costs, or incomplete training attempts as completed training units.

\begin{figure*}[!t]
\centering
\includegraphics[width=\textwidth]{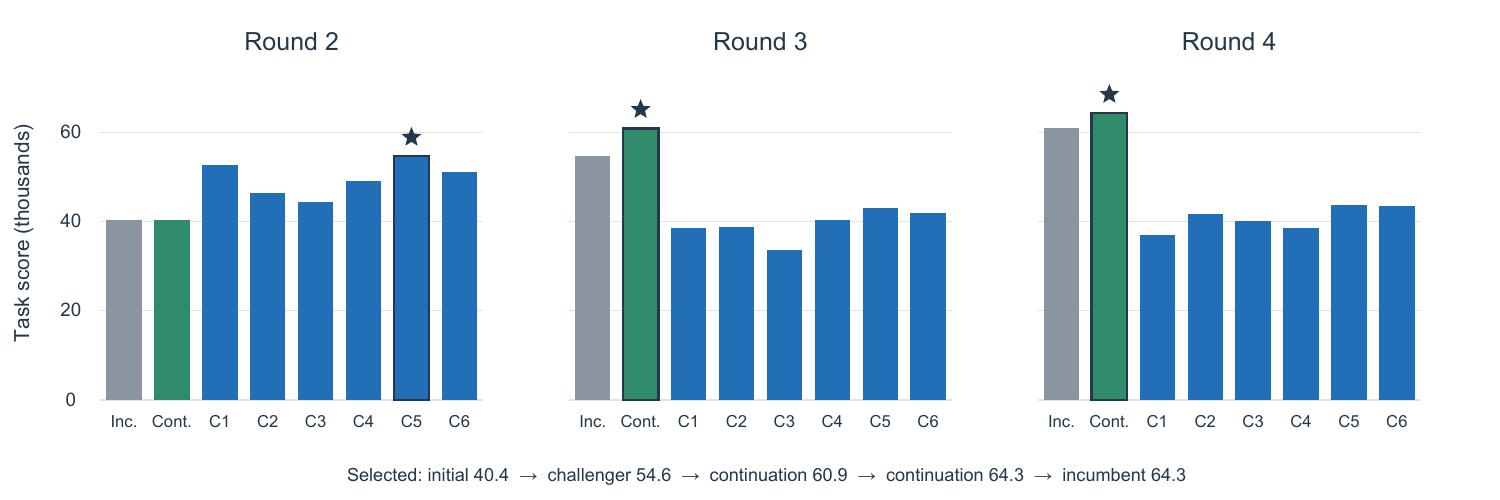}
\caption{Ant candidate scores in rounds 2--4 and the five-round selection path. Stars mark winners. Inc.: incumbent; Cont.: continuation; C1--C6: challengers ordered by identifier within each round.}
\label{fig:supp-ant-selection}
\end{figure*}

\begin{table}[htb]
\centering
\caption{Additional Full searches on Ant; 30 generation units per search. Scores are historical selection scores.}
\begin{tabular}{lrr}\hline
Search & RL seed & Selected score\\\hline
1 & 0 & 57,560.79\\
2 & 0 & 55,259.83\\
3 & 0 & 64,252.91\\\hline
\end{tabular}
\end{table}

\begin{figure}[htb]
\centering
\includegraphics[width=\columnwidth]{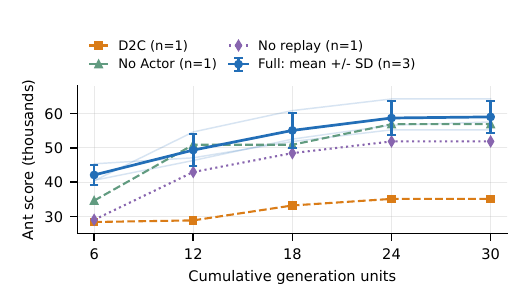}
\caption{Exploratory search-level variation on Ant versus cumulative generation units. Full shows the mean and sample standard deviation of three searches at RL seed 0; other methods have one search each.}
\label{fig:supp-ant-search-variation}
\end{figure}

These three Full searches are not a matched repeated-search comparison with D2C. The 128 episode evaluations within each checkpoint score are rollouts of one policy, not independent proposal--training--selection searches. The main text separately evaluates the final selected policies with three evaluation seeds.

\section{Cross-Iteration Evolution}\label{app:cross-iteration}

\subsection{New-design performance across rounds}
Table~\ref{tab:supp-walker-rounds} compares the highest-scoring newly generated design in each Walker2D round. Round 1 trains the initial population; subsequent LACE rounds train design challengers initialized from the selected Actor. The table excludes the retained incumbent and continuation candidate. Each entry is a within-round maximum, not a cumulative best.

\begin{table}[H]
\centering\footnotesize
\caption{Walker2D: best new design in each round.}
\label{tab:supp-walker-rounds}
\setlength{\tabcolsep}{5pt}
\begin{tabular}{@{}rrrr@{}}\hline
Round & D2C & LACE--CRAFT & Gain\\\hline
1 & 11,023.11 & 12,435.48 & $+12.8\%$\\
2 & 14,422.04 & 16,488.11 & $+14.3\%$\\
3 & 14,902.87 & 16,742.91 & $+12.3\%$\\
4 & 13,902.85 & 15,088.00 & $+8.5\%$\\
5 & 13,186.34 & 16,394.60 & $+24.3\%$\\\hline
\end{tabular}
\end{table}

LACE--CRAFT produces a higher-scoring new design in all five rounds. Its round-wise maximum decreases in round 4 before recovering in round 5, whereas incumbent retention keeps the selected score from decreasing. The final selected Walker2D score is 18,483.50 because selection also admits continuation. These independently generated searches compare the complete methods; their differences do not isolate Actor inheritance.

\subsection{Continuation and challenger selection on Ant}
Figure~\ref{fig:supp-ant-selection} shows all eight comparison states in Ant rounds 2--4: the retained incumbent, one continuation, and six challengers. The initial population selects a robot scoring 40,420.20. In round 2, a design challenger reaches 54,646.39, while continuation scores 40,341.80. Continuation then wins rounds 3 and 4 with 60,870.41 and 64,252.91. Round 5 retains the incumbent at 64,252.91; its additional continuation scores 63,058.92.

The round-2 challenger improves the incoming incumbent by 35.2\%; continuation adds 11.4\% in round 3 and 5.6\% in round 4. Thus design changes and further training both contribute to this search. Across five rounds, LACE--CRAFT's cumulative best increases from 40,420.20 to 64,252.91; D2C's increases from 28,408.55 to 35,139.39. These are recorded outcomes of individual searches, rather than estimates of how often each branch wins across seeds.

\section{Ablation Configurations and Results}\label{app:ablation-config}

\subsection{Variant definitions}
The ablation study compares the full method, No Actor inheritance, and No replay evidence on Ant, HalfCheetah, Hopper, Swimmer, and Walker2D. All retain Feedback, Morphology, Reward, Integration, and reciprocal cross-review. Full and both ablations share the same five-round schedule: 30 new morphology--reward pairs and four full-state continuation updates, with the same per-branch training budget. Each search therefore uses 34 training units.

No Actor inheritance uses fresh challenger Actors and fresh observation-normalization statistics. Value or critic parameters, optimizers, collection state, and replay buffers are initialized as for a new learner. The variant changes both the learned action mapping and its observation scaling. 

No replay evidence excludes replay frames, trajectory-level evidence, and their derived behavioral diagnoses from all role inputs. It retains the current-robot binding, scalar evaluation scores, candidate rankings, and design history. Feedback remains active with these retained inputs.

\begin{table}[H]
\centering\footnotesize
\caption{Historical search-selection scores and completed training budgets. These are not the post-selection three-seed means in the main text. Units include continuations.}
\label{tab:supp-ablation-results}
\setlength{\tabcolsep}{3pt}
\begin{tabular}{@{}llrrr@{}}\hline
Task & Variant & Score & Units & Cont.\\\hline
Ant & Full & 64,252.91 & 34 & 4\\
 & No Actor & 56,901.25 & 34 & 4\\
 & No replay & 51,878.46 & 34 & 4\\[2pt]
HalfCheetah & Full & 27,458.55 & 34 & 4\\
 & No Actor & 25,263.65 & 34 & 4\\
 & No replay & 25,277.06 & 34 & 4\\[2pt]
Hopper & Full & 7,124.44 & 34 & 4\\
 & No Actor & 6,595.88 & 34 & 4\\
 & No replay & 6,693.28 & 34 & 4\\[2pt]
Swimmer & Full & 9,789.24 & 34 & 4\\
 & No Actor & 4,254.11 & 34 & 4\\
 & No replay & 4,835.91 & 34 & 4\\[2pt]
Walker2D & Full & 18,483.50 & 34 & 4\\
 & No Actor & 10,899.12 & 34 & 4\\
 & No replay & 13,342.64 & 34 & 4\\\hline
\end{tabular}
\end{table}

\subsection{Ant round-wise outcomes}
The Ant No Actor variant reaches selected scores of 34,631.61, 50,914.84, 50,914.84, 56,901.25, and 56,901.25 across five rounds. No replay reaches 51,878.46 in round 4 and retains this score in round 5. In the full Ant search, a design challenger wins round 2, continuation wins rounds 3 and 4, and round 5 retains the incumbent.

\begin{table*}[!t]
\centering\footnotesize
\caption{Selected scores after each ablation round. Repeated values indicate retention of the best available robot.}
\label{tab:supp-ablation-rounds}
\setlength{\tabcolsep}{7pt}
\begin{tabular}{@{}llrrrrr@{}}\hline
Task & Variant & Round 1 & Round 2 & Round 3 & Round 4 & Round 5\\\hline
Ant & No Actor & 34,631.61 & 50,914.84 & 50,914.84 & 56,901.25 & 56,901.25\\
 & No replay & 29,049.02 & 42,973.31 & 48,472.22 & 51,878.46 & 51,878.46\\[2pt]
HalfCheetah & No Actor & 25,263.65 & 25,263.65 & 25,263.65 & 25,263.65 & 25,263.65\\
 & No replay & 20,733.75 & 23,762.41 & 24,715.28 & 25,277.06 & 25,277.06\\[2pt]
Hopper & No Actor & 5,212.28 & 6,595.88 & 6,595.88 & 6,595.88 & 6,595.88\\
 & No replay & 6,573.45 & 6,573.45 & 6,573.45 & 6,573.45 & 6,693.28\\[2pt]
Swimmer & No Actor & 2,876.15 & 2,876.15 & 2,876.15 & 2,876.15 & 4,254.11\\
 & No replay & 2,466.44 & 2,718.34 & 2,718.34 & 2,718.34 & 4,835.91\\[2pt]
Walker2D & No Actor & 9,239.89 & 9,890.62 & 9,890.62 & 10,899.12 & 10,899.12\\
 & No replay & 13,342.64 & 13,342.64 & 13,342.64 & 13,342.64 & 13,342.64\\\hline
\end{tabular}
\end{table*}

\subsection{Scope of the comparisons}
Each variant generates its own initial candidates. Differences can therefore arise before the first transfer step and propagate through later proposals, training, and selection. Because No Actor resets both the Actor and observation statistics, the contrast concerns their joint inheritance under the common continuation schedule. The full-search initialization settings are listed in Table~\ref{tab:supp-five-init}. No replay uses the same unscaled Actor/normalizer copying and full-state continuation as Full; only the replay evidence is removed.

The replay comparison yields higher full-method scores on all five tasks. This comparison assesses replay evidence within the retained four-role collaboration; it does not isolate the contribution of the collaboration structure itself. The historical scores in this section are individual search outcomes, without an estimate of variability across independent search seeds. The main ablation table separately reports fixed-policy evaluation variability over three evaluation seeds.

\section{Generative Geometry and Hardware Interfaces}\label{app:generative-interface}

\subsection{Artifacts at each processing stage}
The visual reference supplies a shape cue for Rodin mesh generation. BANG separates the generated surface into body and limb meshes; repair, scaling, and alignment establish the link geometry used for joint placement. Motor cavities and mating interfaces are then fitted before compilation of the articulated model.

\begin{table}[H]
\centering\footnotesize
\vspace*{7pt}
\caption{Artifacts produced by generative-morphology processing.}
\label{tab:supp-generative-artifacts}
\begin{tabular}{@{}p{0.28\columnwidth}p{0.64\columnwidth}@{}}\hline
Stage & Output\\\hline
Shape generation & Static mesh conditioned on the morphology reference.\\[3pt]
Decomposition and alignment & Separate link meshes $g_i$ with local frames and aligned connections.\\[3pt]
Joint allocation & Parent--child graph, joint origins, types, axes, and position limits.\\[3pt]
Actuator fitting & Motor positions and orientations, cavity geometry, and mating surfaces in $\Sigma$.\\[3pt]
Compilation & URDF, collision geometry, inertial properties, joint and actuator limits, and action mapping in $\Phi$.\\\hline
\end{tabular}
\end{table}

\subsection{Dependencies of a morphology edit}
A link edit changes geometry within $G_m$. Affected joint origins and motor placements must remain consistent with the revised shape and local coordinate frames. Cavity dimensions are constrained by the selected motors; mating surfaces must remain aligned between the connected links. Collision geometry, masses, centers of mass, and inertia tensors are updated with the compiled model.

The action mapping $\Gamma$ binds policy outputs to an ordered set of actuated joints and defines their scaling and control frequency. Fixed observation and action dimensions and consistent joint ordering are required for direct Actor inheritance. Changes in articulation that alter these interfaces require a compatible controller specification before training.

Hardware attachment remains partly manual. Screw-hole placement and mating details are designed for the selected components. The printed parts were fabricated without prior structural simulation, and thin connections were prone to fracture. Structural analysis and load-aware redesign remain necessary for more reliable fabrication.

\subsection{Raspberry Pi controller and physical recording}
The indoor walking recording uses the R17 hardware-adapted PPO policy at training epoch 650. It was trained with the Isaac Gym robot task and exported from the PyTorch checkpoint to a NumPy parameter archive. This controller belongs to the generative-robot hardware adaptation, separately from the five Brax benchmark policies. Its hardware-specific training uses three actuated joints per leg (12 actions), whereas the illustrated geometry-stage model uses two per leg. The exported actor has hidden widths 512, 256, and 128 with ELU activations and a 12-dimensional output. The recorded export check reports a maximum absolute NumPy--PyTorch output difference of $4.17\times10^{-7}$.

Policy inference runs locally on the Raspberry Pi at 20 Hz. The 58-dimensional observation contains 12 measured joint positions, 12 joint velocities estimated by one-step finite differences, 12 previous actions, eight sinusoidal gait-clock signals, three command entries, and 11 fixed placeholder entries. The placeholders comprise unavailable base linear and angular velocities, a fixed gravity direction, and unavailable planar position-error signals. No IMU is required. Observation normalization statistics are exported with the actor; normalized inputs are clipped to $[-5,5]$. The runtime reproduces the one-control-frame observation delay used in the adaptation.

Actions are clipped to $[-1,1]$, low-pass filtered with coefficient approximately $0.49646$, and scaled by $0.63$ radians about the reference pose. Joint-target changes are limited to $0.145$ radians per control step and clipped to joint limits. Calibrated zero offsets, signs, and radians-per-unit mappings convert targets into servo-bus commands. The policy joint ordering maps to servo IDs $1,5,9,2,6,10,3,7,11,4,8,12$. The internal gait clock runs at $1.05$ Hz. These settings describe the R17 deployment; the arm is absent during the recorded walking trial.

The figure uses frames at 24, 28, 32, and 36 seconds from one continuous indoor recording. Equal-sized crops follow the robot to keep its body and feet visible. They document physical motion rather than calibrated displacement or benchmark task scores. Quantitative comparisons in the main tables concern the simulated benchmark tasks.

\subsection{Manipulation interface}
A mount is incorporated into the base link for an SO-101 arm, while the arm's existing links, motors, and gripper are reused. The assembled photograph shows the arm installed; the walking recording shows the platform without it. Simulation controls the leg joints while holding the arm at prescribed fixed postures. Initial manipulation trials performed poorly. Effective mobile manipulation requires additional control and mechanical validation. Measured joint motion and real-world video could subsequently supply blackboard feedback.

\section{Single-search Generation Progress}
\begin{figure}[!htb]
\centering
\includegraphics[width=\columnwidth]{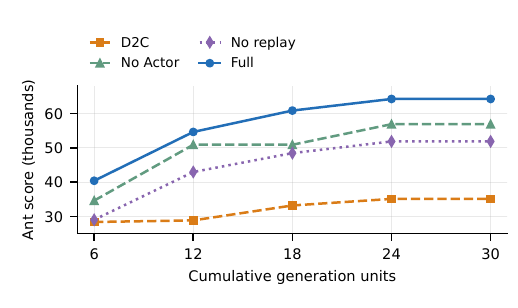}
\caption{Ant single-search progress versus cumulative generation units. Each curve shows stored selection scores from one search, not the post-selection evaluation means.}
\label{fig:budget-variability}
\end{figure}

Figure~\ref{fig:budget-variability} tracks the reported Ant search for each method after $6,12,18,24,30$ cumulative generation units. Each point is the selected checkpoint score at the end of that round. The Full curve ends at the stored selection score $64{,}252.91$; the main tables report a separate three-seed evaluation of the selected policy. Each curve represents one complete search.

\end{document}